\documentclass[a4paper]{article}

\usepackage[utf8]{inputenc}
\usepackage{erk}
\usepackage{times}
\usepackage{graphicx}
\usepackage[top=22.5mm, bottom=22.5mm, left=22.5mm, right=22.5mm]{geometry}
\usepackage{multirow}
\usepackage{color, colortbl}
\usepackage[breaklinks=true,letterpaper=true,colorlinks,bookmarks=false]{hyperref}
\usepackage[english]{babel}
\usepackage[dvipsnames]{xcolor}

\def\footnotemark{}%  to avoid footnote on cover page

\begin{document}
%make title
\title{Face Morphing Attack Detection Using \\Privacy-Aware Training Data\thanks{Supported in parts by the ARRS project J2-1734 (B), and the ARRS research programmes P2-0250 (B) and P2-0214 (B).}}

\author{Marija Ivanovska$^{1}$, Andrej Kronov\v{s}ek$^{2}$,  Peter Peer$^{2}$, Vitomir \v{S}truc$^{1}$, Borut Batagelj$^{2}$} % use ^1, ^2 for author(s) from different institutions

\affiliation{$^{1}$Faculty of Electrical Engineering, University of Ljubljana, Tržaška cesta 25, SI-1000 Ljubljana, Slovenia\\
$^{2}$Faculty of Computer and Information Science, University of Ljubljana, Večna pot 113, SI-1000 Ljubljana, Slovenia}

\email{E-mail: marija.ivanovska@fe.uni-lj.si}

\maketitle

\begin{abstract}{Abstract}
Images of morphed faces pose a serious threat to face recognition--based security systems, as they can be used to illegally verify the identity of multiple people with a single morphed image. Modern detection algorithms learn to identify such morphing attacks using authentic images %examples of bona fide and attack images 
of real individuals. This approach raises various privacy concerns and limits the amount of publicly available training data. %To alleviate the problem, 
In this paper, we explore the efficacy of detection algorithms that are trained only on faces of non--existing people and their respective morphs. To this end, two dedicated algorithms are trained with synthetic data and then evaluated on three real-world datasets, i.e.: FRLL-Morphs, FERET-Morphs and FRGC-Morphs. Our results show that synthetic facial images can be successfully employed for the training process of the detection algorithms and generalize well to real-world scenarios. 
\end{abstract}

\selectlanguage{english}
\vspace{-3 pt}
\section{Introduction}
\vspace{-3 pt}

Nowadays, the vast majority of applications for person identity verification rely on Face Recognition Systems (FRSs), which match a human face to an entry from a database of faces. Modern FRSs have proven to be highly accurate when genuine faces are presented to the system \cite{grm2018strengths}. They are however prone to various attacks, whose aim is to gain illegal access by false authentication \cite{Huber_SYN-MAD_2022}. %3D face masks, electronic display images, printed face images, just to name a few, are all frequently used to subvert the FRSs. 

Lately, face morphs have become a growing concern for the reliability of face verification systems. A face mor\-ph is a composite image generated from two (or more) facial images of distinct subjects. % and resemble an authentic subject. 
Recent advances in generative deep models have enabled an almost effortless generation of realistic and high-quality morphed facial images. Such images can be utilized to verify all identities that have been used in the morph-generation process. A successful detection of \textit{face morphing attacks} is therefore critical for the prevention of illegal activities \cite{Naser_PW_MAD_2021}.

\begin{figure}[!t]
\begin{center}
\centering
  \includegraphics[width=0.9\linewidth]{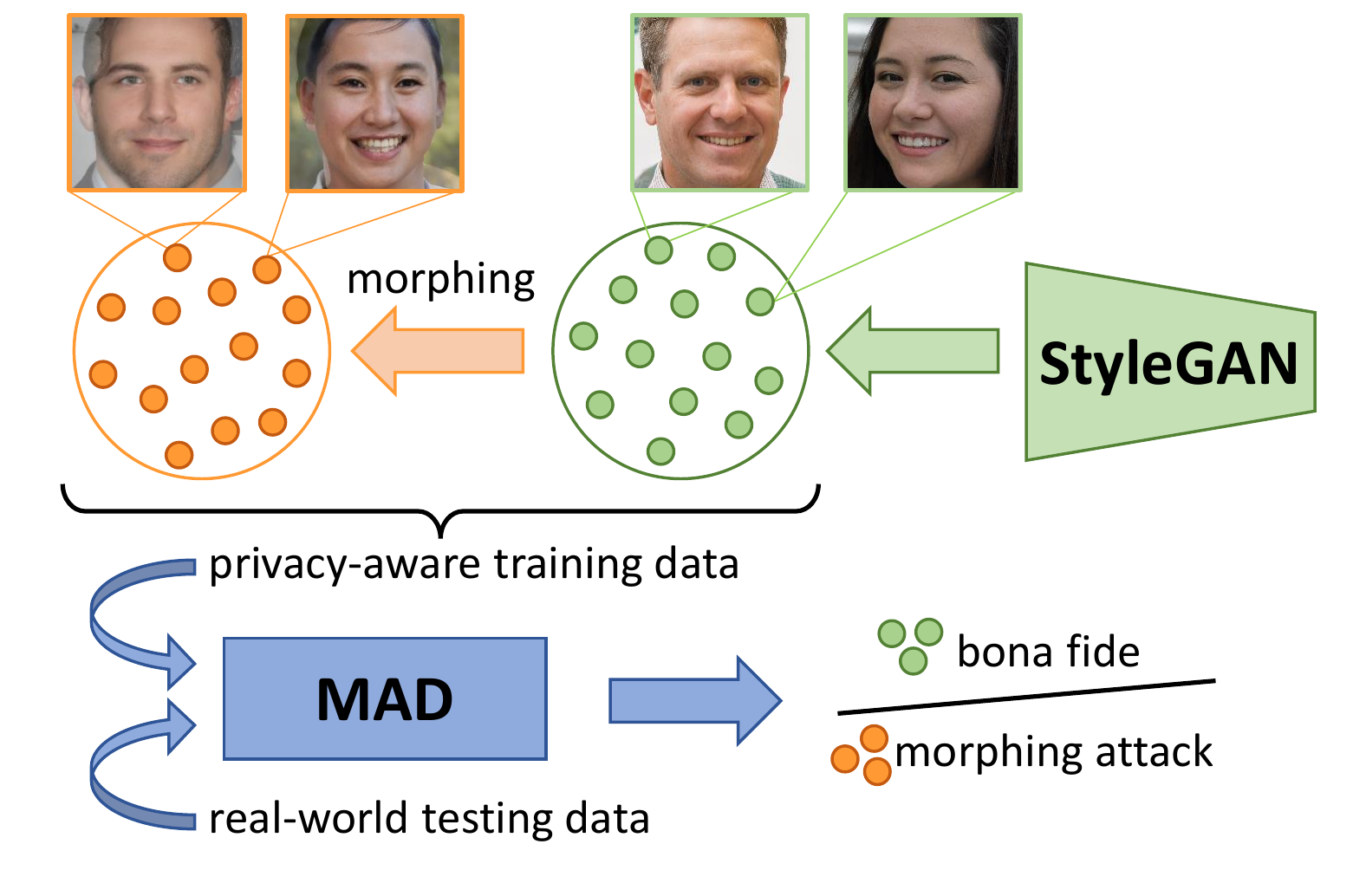}
\end{center}
\vspace{-20pt}
\caption{To avoid privacy related concerns in the development of morphing attack detectors (MAD), we explore the idea of using synthetic training faces of non--existing people. Trained MADs are then tested on real--world datasets.\label{fig:teaser}}
\end{figure}
% \vspace{-10pt}

Various \textit{Morphing Attack Detection} (MAD) algori\-thms have been proposed over the years to automatically distinguish real from morphed faces. 
%Compared to real facial images, face morphs often contain different artifacts, generated during the morphing process. Existing state--of--the--art MAD approaches utilize these traces in a discriminative manner, by analyzing different image features such as the texture, the image quality, the presence of residual noise etc. Regardless of the detection technique, the training of modern MAD models requires a massive database of genuine face images. 
However, regardless of the detection technique, the training of these models requires a massive database of genuine face images. Training protocols therefore raise various privacy-related concerns and limit the amount of publicly available training data that can be used to learn MAD models. In this paper, we address the privacy issues related to MADs by exploring the idea of using synthetic training data, as illustrated in Figure~\ref{fig:teaser}. For this purpose, we use the SMDD~\cite{Damer_SMDD_2022} dataset, where StyleGAN2 \cite{karras2020analyzing}, a state-of-the-art Generative Adversarial Network (GAN), was utilized to generate \textit{bona fide} face images of non-existing people. These images were then used for the generation of the face morphs. %, with commonly used face morphing techniques. 
With this dataset, we train two powerful binary classifiers, Xception and HRNet, and evaluate their detection performance on three real-world datasets -- FRLL-Morphs, FERET-Morphs and FRGC-Morphs. The results of the evaluation show that well-performing MAD models can be learned from synthetic data alone, and that the model generalize well over three diverse real-world morph datasets.   
\vspace{-3 pt}
\section{Related work}
\vspace{-3 pt}
Existing morphing-attack-detection models can in general be categorized as single--image (S-MAD) or differential (D-MAD) MADs, depending on whether the face morph is examined independently or is compared to a reference sample. While D-MADs can be very accurate in closed--set problems, S-MADs aim to detect attacks without any prior knowledge about human identities. In this section, we only review S-MADs, since they are more closely related to our work. 

% \subsection{Hand--crafted MADs} 
Regardless of the face morphing technique used, the generated morphs usually contain image irregularities su\-ch as artifacts, noise, pixel discontinuity, distortions, spectrum discrepancies, inconsistent illumination, etc. In the past, shallow algorithms, that implement extraction of photo-response non-uniformity (PRNU) noise~\cite{Scherhag_PRNU_2019} or reflection analysis~\cite{Seibold_reflection_2018} have been successfully employed for the detection of morphing attacks. Some other hand--crafted MAD methods have also used texture--based descriptors, such as LBP~\cite{Ojala_1996_LBP}, LPQ~\cite{Ojansivu_LPQ_2008} or SURF%~\cite{Kraetzer_SURF_2017
~\cite{Makrushin_SURF_2019}. Although these methods achieved promising results, they were shown to have limited generalization capabilities. Moreover, as the face morphing techniques improved over time, the performance of shallow methods became less competitive, as they struggled to detect modern, deep-learning generated or heavily post--processed face morphs.

% \subsection{Deeplearning MADs} 
More recent MAD models take advantage of the development of data--driven, deep-learning algorithms. Rag\-havendra \textit{et al.}~\cite{Raghavendra_transfer_2017} were amongst the first to propose transfer learning. In their work, attacks are detected with a simple, fully--connected binary classifier, fed with fused VGG19 and AlexNet features, pretrained on ImageNet. Wandzik \textit{et al.}~\cite{Wandzik_FRS_2018}, on the other hand, achieve highly accurate results with features from general--purpose face recognition systems (FRSs) combined with an SVM. Ramachandra \textit{et al.}~\cite{Ramachandra_Inception_2020} utilize Inception in a similar manner, while Damer \textit{et al.}~\cite{Naser_PW_MAD_2021} argue, that pixel--wise supervision, where each pixel is classified as a bona fide or a morphing attack, is superior, when used in addition to the binary, image--level objective. % Their PW-MAD is based on the DenseNet architecture. 
Recently, MixFaceNet~\cite{Boutros_MixFaceNet_2021} by Boutros \textit{et al.} achieved state--of--the--art results in different face--related detection tasks, including face morphing detection~\cite{Damer_SMDD_2022}. This model represents a highly efficient architecture that captures different levels of face attack cues by using differently sized convolutional kernels.
\vspace{-3 pt}
\section{Methods}
\vspace{-3 pt}
We consider two different classification models, Xception and HRNet, to detect face morphing attacks in this study and train them using synthetic data only. The two models represent the entries from the University of Ljubljana to the recent \textit{Face Morphing Attack Detection Competition based on Privacy-aware Synthetic
Training Data} (SYN-MAD), held in conjunction with the 2022 International Joint Conference on Biometrics (IJCB 2022)~\cite{Huber_SYN-MAD_2022}, which achieved the best and third best overall performance among all submitted entries. %Details on the two models are given below.    

\textbf{Xception}~\cite{Xception} is a convolutional neural network (CNN) that updates and simplifies the architecture of the InceptionV3 model \cite{szegedy_rethinking_2015} by replacing the Inception modules with depth--wise separable convolutions. %Moreover, authors of Xception don’t use the ReLU activations after the convolutional layers and add residual connections, that were shown to help with the convergence of the model. 
We use Xception as a feature extractor, while the binary classification is performed by a fully connected two--layer network. The output layer consist of $2$ neurons, followed by a softmax activation function. %Such architecture is a commonly used baseline for face morphing attacks detection. 
Similar to previous research, we use cross--entropy as the learning objective.
% $$\mathcal{L}_{ce} =  -\sum_{i=1}^{2} y_i \log(p(y_i)).$$

\textbf{HRNet}~\cite{Wang_HRNet_2019} is again a CNN that unlike other networks maintains high--resolution representations of the input sample through the whole feedforward process. %High--to--low resolution convolutions are connected in parallel, by fusing them together at each downsampling layer. 
Su\-ch an architecture contributes to more descriptive image representations, which was proven to improve the results of different classification tasks. %Recently, HRNet was successfully employed for biometric data, as a backbone network for deepfake detection. 
In our experiments, we replace the classification head of HRNet with a two--layer classification module, to perform binary detection of bona fide images and morphing attacks. The output layer consists of only one neuron, followed by a sigmoid activation function. In the training phase, the parameters are optimized using the binary cross--entropy loss.
% $$\mathcal{L}_{bce} =  -\big[\log(p(y_i)) + \log(1-p(y_i))\big].$$
\begin{figure}[t]
\begin{center}
\centering
  \includegraphics[width=\linewidth]{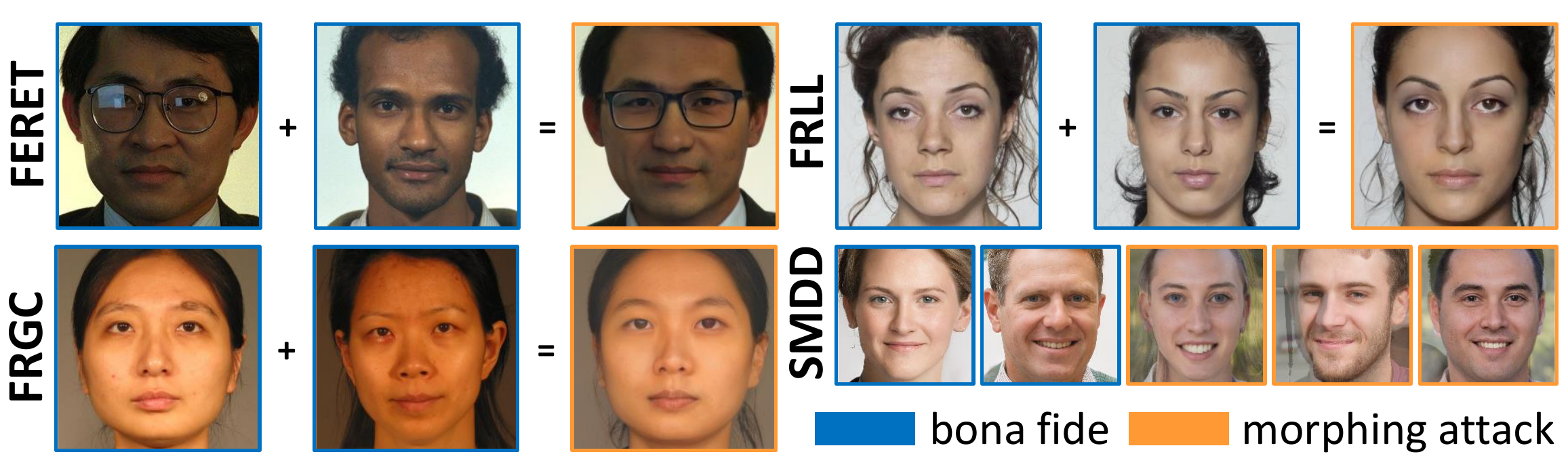}
\end{center}
\vspace{-20pt}
\caption{Examples of bona fide and morphing attack images from the SMDD~\cite{Damer_SMDD_2022} training dataset and the testing datasets FERET-Morphs, FRLL-Morphs and FRGC-Morphs~\cite{Sarkar2020_morphed_data}. \label{fig:datasets}}
\end{figure}
\begin{table}[t]
\caption{Number of bona fide images (BF), number of morphing attacks (MA) generated by morphing methods OpenCV (OCV), FaceMorpher (FM), StyleGAN (SG), AMSL, Webmorpf (WM) and image size of samples in each dataset.\vspace{-4mm}}
\label{tab1}
\begin{center}
\resizebox{\columnwidth}{!}{%
\begin{tabular}{ | l | c | c | c | c | c | c | c |}
\hline  
  \textbf{Dataset} & \textbf{Image size} & \textbf{BF} & \textbf{OCV} & \textbf{FM} & \textbf{SG} & \textbf{AMSL} & \textbf{WM} \\ 
\hline  
%   SMDD & $256\times256$ & $25.000$ & $15.000$\\
  FRLL-M & $1350\times1350$ & $204$ & $1221$ & 1222 & 1222 & 2175 & 1221 \\ %$7.061$\\
  FERET-M & $512\times768$ & $1.413$ & 529 & 529 & 529 & / & / \\%$1.587$\\
  FRGC-M & $227\times277$ & $3.167$ & 964 & 964 & 964 & / & / \\%$2.892$\\
\hline 
\end{tabular}}\vspace{-4mm} 
\end{center}
\end{table}

\vspace{-3 pt}
\section{Experiments}
\vspace{-3 pt}
\subsection{Datasets}
We use one synthetic and three publicly available real--world datasets in this work. The training is done exclusively with the synthetic data, while the evaluation is performed on three commonly used face morphing datasets.

\textbf{Training data.}
For training, we use the SMDD data\-set \cite{Damer_SMDD_2022}, provided by the organizers of the SYN-MAD competition~\cite{Huber_SYN-MAD_2022}. The dataset consists of $25.000$ bona fide and $15.000$ morphed images of size $256\times256$ pixels. 
Bona fide instances represent carefully selected images from a set of randomly generated StyleGAN2~\cite{karras2020analyzing,Karras_stylegan2_2020} faces. A separate, non -- overlapping StyleGAN2 image set was used for the generation of face morphing attacks. Face morphs were created using the landmark--based morphing technique from OpenCV\footnote{https://learnopencv.com/face-morph-using-opencv-cpp-python/}. A few selected samples from the SMDD dataset are presented in Figure~\ref{fig:datasets}.
% to se uporablja samo za generiranje evaluacijske baze: and two different deeplearning--based methods, i.e. MIPGAN-I and MIPGAN-II~\cite{Zhang_MIPGAN_2021}. 
\begin{figure*}[!t]
\begin{center}
\centering
  \includegraphics[width=0.345\linewidth]{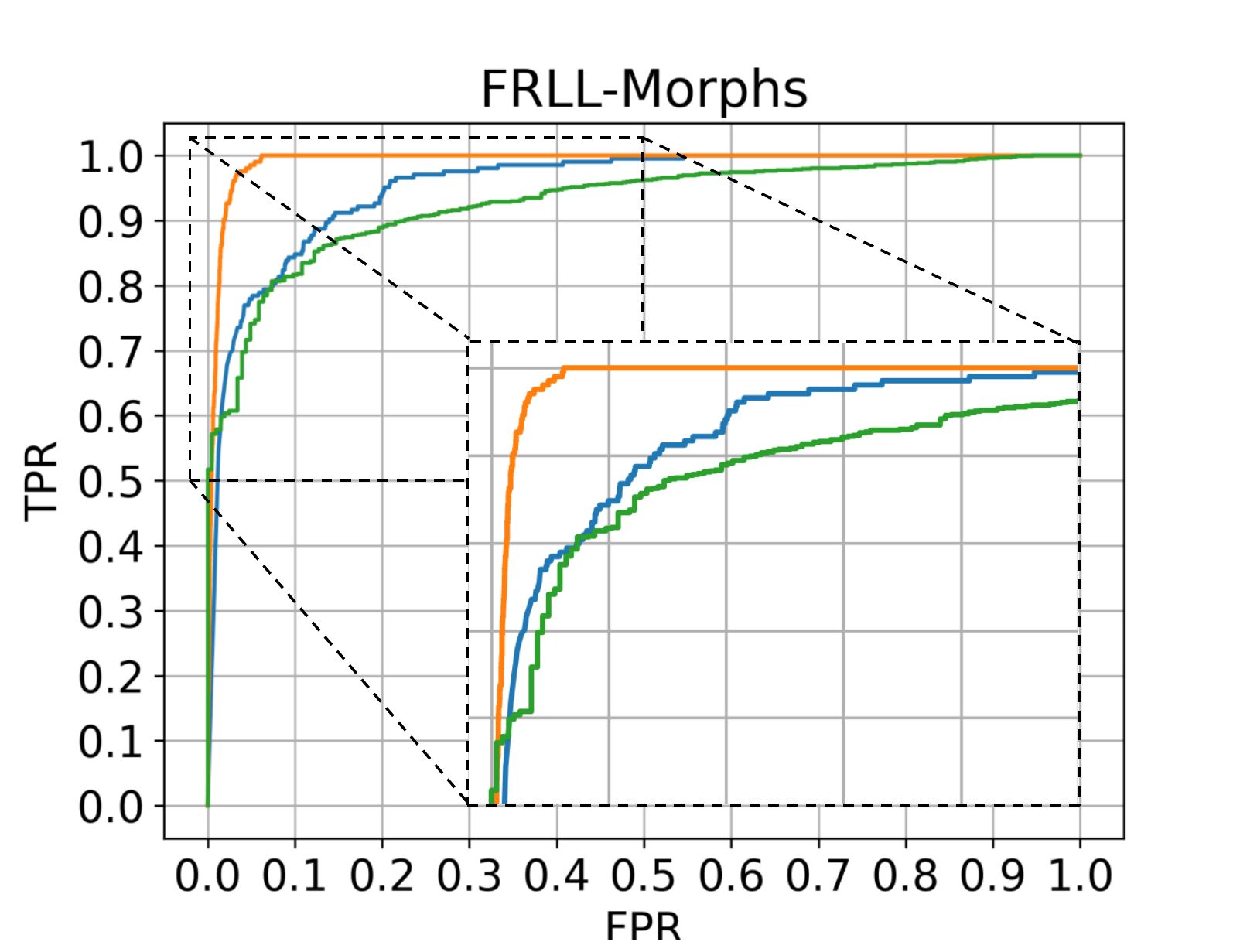} \hspace{-15 pt}
  \includegraphics[width=0.345\linewidth]{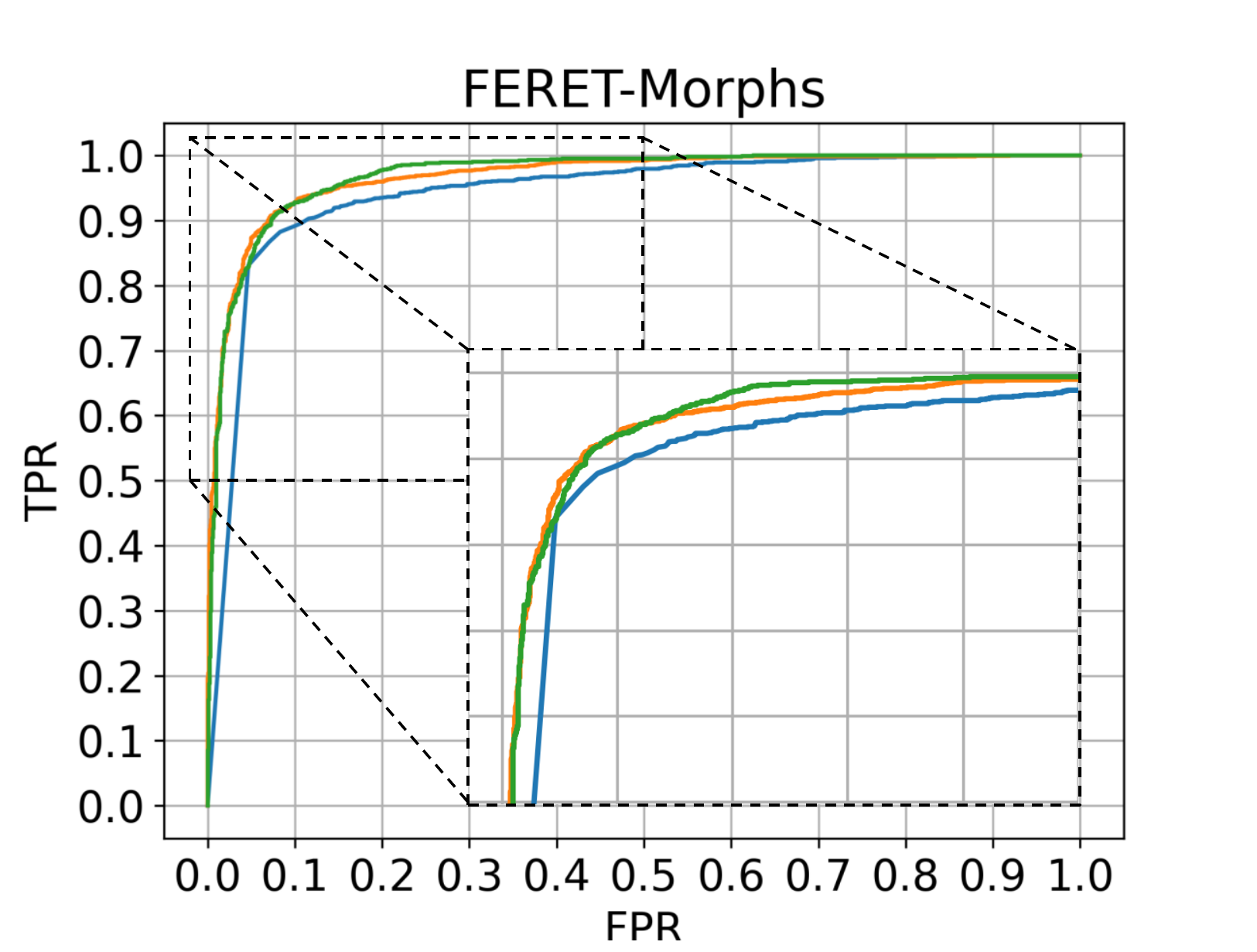}\hspace{-15 pt}
  \includegraphics[width=0.345\linewidth]{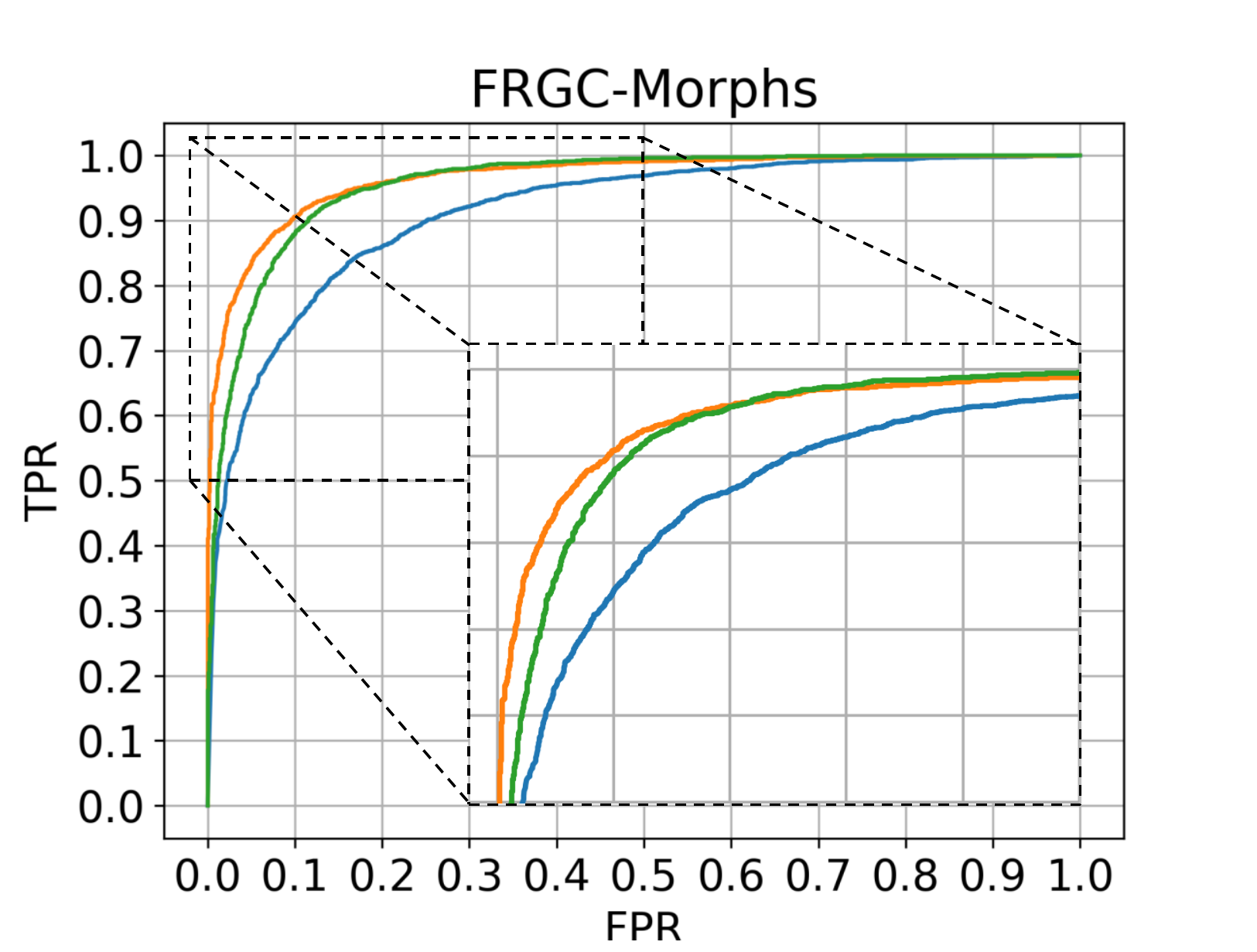}
  \includegraphics[width=0.345\linewidth,trim = 0mm 100mm 0 0, clip]{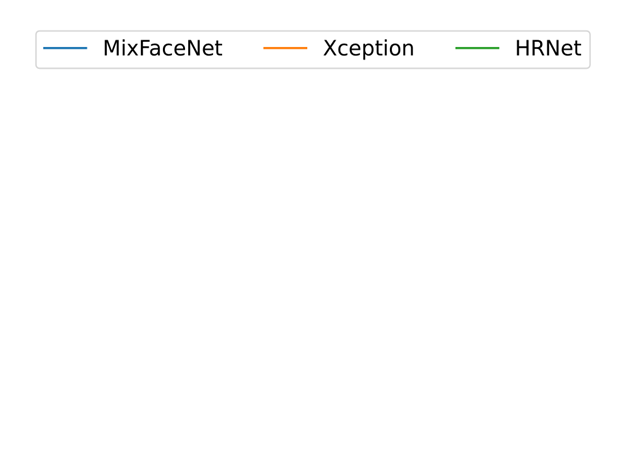}
\end{center}
\vspace{-25 pt}
\caption{ROC curves generated on FRLL-Morphs, FERET-Morphs and FRGC-Morphs for the tested models. Note that HRNet achieves very competitive results on FERET-Morphs and FRGC-Morphs, but performs the weakest on FRLL-Morphs. Xception on the other hand, consistently outperforms the baseline method MixFaceNet, on all considered datasets. \label{fig:ROC_by_datasets}}
\end{figure*}

%\vspace{-5 pt}
\textbf{Testing data.}
The trained MAD models are tested on three diverse morphing datasets proposed by Sarkar \textit{et al.} in~\cite{Sarkar2020_morphed_data}, i.e. FRLL-
Morphs, FERET-Morphs and FRGC-Morphs. All face morphs were created by combining bona fide images from their respective face datasets, i.e. FRLL~\cite{debruine_jones_2017, Neubert_FRLL}, FERET~\cite{PHILLIPS_FERET} and FRGC~\cite{Phillips_FRGC}. To generate landmark--based morphs, the authors used OpenCV and FaceMorpher\footnote{https://github.com/alyssaq/face\_morpher}, while deep-learning--based morphs are generated with StyleGAN2. In addition to these methods, AMSL\cite{Neubert_FRLL} and Webmorph\footnote{https://webmorph.org/} are also used, but only for the images from the FRLL dataset. Information about image sizes and the number of samples per morphing method is given in Table~\ref{tab1}. Selected samples from all three datasets are presented in Figure~\ref{fig:datasets}. 
%\vspace{-15 pt}

\vspace{-5 pt}
\subsection{Experimental setup}
In our experiments, we first preprocess images from all datasets by cropping out the facial areas. Bounding boxes of the SMDD are provided by the authors of the dataset. For the other three databases, we use RetinaFace~\cite{RetinaFace}, to localize the facial region-of-interest. Prior to their use, cropped images are resized to $299\times299$ pixels for Xception and $256\times256$ pixels for HRNet. Additionally, the training data is augmented with horizontal flips to increase the amount of data available and avoid overfitting. % and random rotations for up to $5$ degrees.

The CNNs were optimized using the Adam optimizer, with a learning rate of $0.0001$. The models were trained from scratch for $30$ full epochs, with a batch size of $16$. After each training epoch, the classification accuracy of the networks was calculated on a small holdout set of each test dataset. The best performing parameters on each of the three datasets were saved as the final model for that particular dataset. %The training is performed on SMDD samples, while FRLL-Morphs, FERET-Morphs and FRGC-Morphs are used in the testing phase. 
The code was implemented in Python 3.8 with PyTorch 1.9 and CUDA 11.6. Experiments were run on a single GeForce GTX 1080 Ti. The computational complexity of Xception is $11$ GFLOPs, while HRNet has $34$ GFLOPs.%With this hardware Xception needed appx. XXms and HRNet XXms for the detection process on a single image.  

\vspace{-3 pt}
\section{Results}
\vspace{-3 pt}
In Figure~\ref{fig:ROC_by_datasets} and Table~\ref{tab:quantitative_results} we present the results obtained with our two models, Xception and HRNet, and the baseline MixFaceNet-MAD from~\cite{Huber_SYN-MAD_2022}. The weights of MixFa\-ceNet-MAD, optimized on the SMDD dataset, were provided by the authors of the model. In our experiments, the best overall results were achieved by Xception, whose Equal Error Rates (EER) are $3.26\%$, $8.25\%$ and $9.75\%$ on FRRL--Morphs, FERET-Morphs and FRGC--Morphs, respectively (Table~\ref{tab:quantitative_results}). The runner-up, HRNet, achieves a similar performance on FERET-Morphs and FRGC-Mo\-rphs. However, among the tested models, HRNet is least successful on FRLL-Morphs, where it achieves an EER of $13.73\%$. On this dataset, MixFaceNet yields a slightly better performance than HRNet with an EER of $12.18\%$, but is outperformed by both, Xception and HRNet, on the other two databases, i.e. FERET-Morphs and FRGC-Morphs. The complete ROC curves of the experiments are visualized in Figure~\ref{fig:ROC_by_datasets} and show a similar picture as the discussed numerical results.  
\begin{table}[t]
\caption{Detection results for MAD methods MixFaceNet (MFN), Xception (XN) and HRNet (HRN) on the real--world datasets FRLL-Morphs (FRLL-M), FERET-Morphs (FERET-M) and FRGC-Morphs (FRGC-M). Best scores per dataset are marked blue, while runner--up results are marked orange. All three models were trained on the synthetic SMDD dataset \cite{Damer_SMDD_2022}.   %\textcolor{red}{side note: MixFaceNet was trained on full SMDD, but only half of those examples were provided for the competition. Also, Xception was evaluated on a mix of all three datasets, while HRNet is evaluated separately on each dataset.}
}
\vspace{-20pt}
% \vspace{-11 pt}
\label{tab:quantitative_results}
\smallskip
\begin{center}
\resizebox{\columnwidth}{!}{%
\begin{tabular}{ | c | c | c | c | c | c | c | c | }
\hline  
  \multirow{2}{*}{\textbf{MAD}} & \multirow{2}{*}{\textbf{Test data}} & \multirow{2}{*}{\textbf{AUC(\%)}} & \multirow{2}{*}{\textbf{EER(\%)}} & \multicolumn{4}{c |}{\textbf{BPCER (\%) @ APCER =}} \\ \cline{5-8}
   &  &  & & $0.10\%$ & $1.00\%$ & $10.00\%$ & $20.00\%$ \\ \hline
\hline  
  \multirow{3}{*}{MFN~\cite{Damer_SMDD_2022}} & FRLL-M & \textcolor{orange}{$\mathbf{95.43}$} & \textcolor{orange}{$\mathbf{12.18}$} & $100.0$ & $100.0$ & \textcolor{orange}{$\mathbf{15.20}$} & \textcolor{orange}{$\mathbf{5.88}$}\\\cline{2-8}
%   \parbox[t]{2mm}{\multirow{4}{*}{\rotatebox[origin=c]{90}{MixFaceNet}}} & SMDD & $256\times256$ & $25.000$ \\
  & FERET-M & $94.27$ & $10.65$ & $100.0$ & $100.0$ & $11.75$ & $6.51$\\\cline{2-8}
  & FRGC-M & $91.42$ & $16.36$ & $100.0$ & $64.86$ & $25.89$ & $14.02$\\\hline \hline
%   & All & $17.25$ & $100.0$ & $100.00$ & $25.13$ & $14.99$\\\hline
  \multirow{3}{*}{XN~\cite{Xception}} & FRLL-M & \color{blue} $\mathbf{99.17}$ & \color{blue} $\mathbf{3.26}$ & \color{blue} $\mathbf{85.29}$ & \color{blue} $\mathbf{28.92}$ & \color{blue} $\mathbf{0.49}$ & \color{blue} $\mathbf{0.0}$\\\cline{2-8}
  & FERET-M & \textcolor{orange}{$\mathbf{96.84}$} & \color{blue} $\mathbf{8.25}$ & \color{blue} $\mathbf{79.62}$ & \color{blue} $\mathbf{43.31}$ & \color{blue} $\mathbf{7.29}$ & \textcolor{orange}{$\mathbf{4.03}$} \\\cline{2-8}
  & FRGC-M & \color{blue} $\mathbf{96.63}$ & \color{blue} $\mathbf{9.75}$ & \color{blue} $\mathbf{58.19}$ & \color{blue} $\mathbf{35.11}$ & \color{blue} $\mathbf{9.44}$ & \color{blue} $\mathbf{4.23}$\\\hline \hline
%   & All & $14.95$ & $92.10$ & $68.58$ & $21.43$ & $11.08$\\
  \multirow{3}{*}{HRN~\cite{Wang_HRNet_2019}} & FRLL-M & $92.79$ & $13.73$ & $100.00$ & \textcolor{orange}{$\mathbf{42.84}$} & $18.65$ & $11.12$ \\\cline{2-8}
  & FERET-M & \color{blue} $\mathbf{97.05}$ & \textcolor{orange}{$\mathbf{8.49}$} & \textcolor{orange}{$\mathbf{91.43}$} & \textcolor{orange}{$\mathbf{54.00}$} & \textcolor{orange}{$7.44$} & \color{blue} $\mathbf{2.27}$\\\cline{2-8}
  & FRGC-M & \textcolor{orange}{$\mathbf{95.77}$} & \textcolor{orange}{$\mathbf{10.89}$} & \textcolor{orange}{$\mathbf{82.26}$} & \textcolor{orange}{$\mathbf{55.50}$} & \textcolor{orange}{ $\mathbf{12.14}$} & \textcolor{orange}{$\mathbf{4.46}$}\\%\cline{2-7}
%   & All & $20.11$ & $88.60$ & $69.57$ & $33.54$ & $20.17$\\ \hline  
\hline
\end{tabular}}\vspace{-3mm}
\end{center}
\end{table}

%\vspace{-15 pt}

\begin{figure*}[!t]
\begin{center}
\centering
   
  \includegraphics[width=0.345\linewidth]{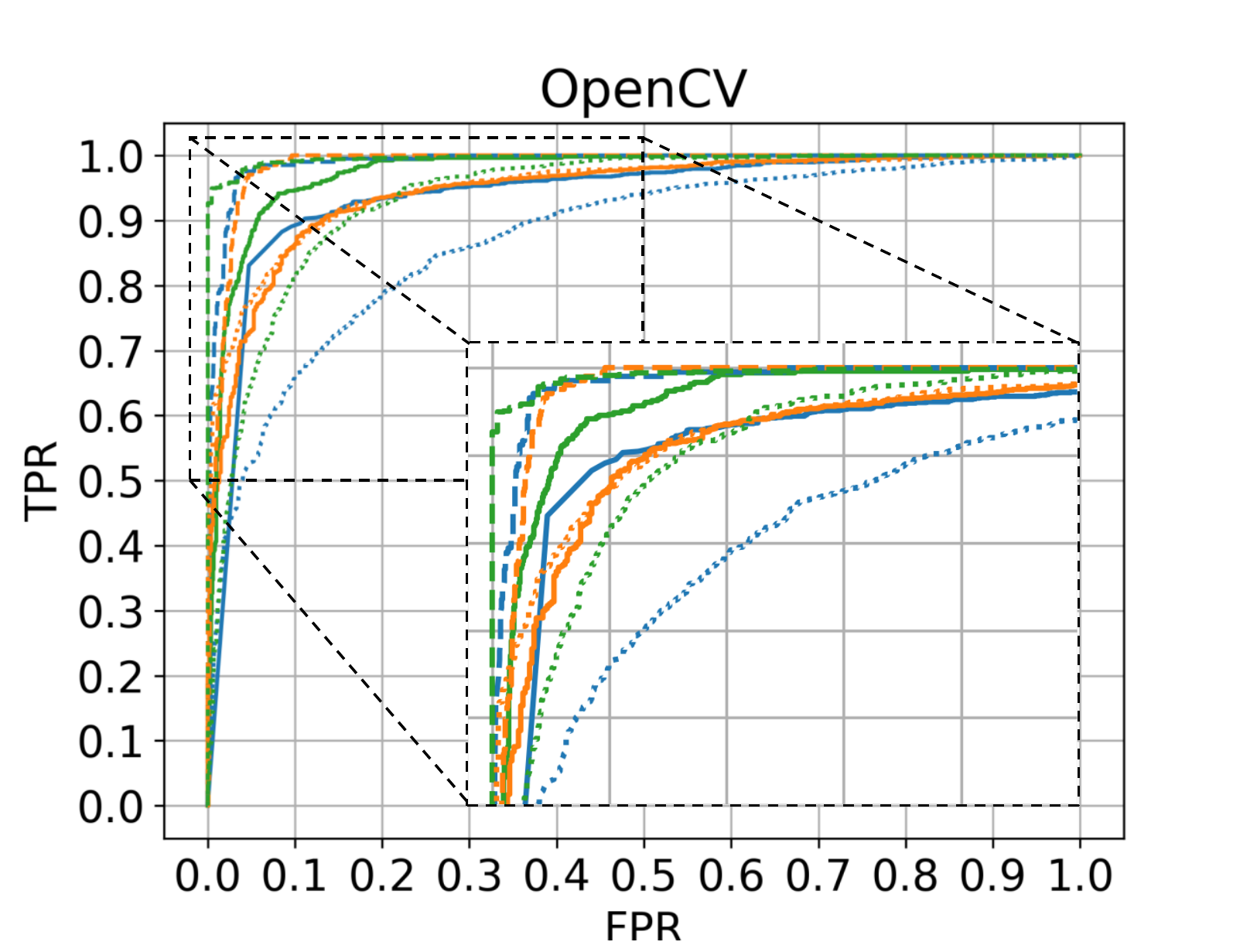}\hspace{-15 pt}
  \includegraphics[width=0.345\linewidth]{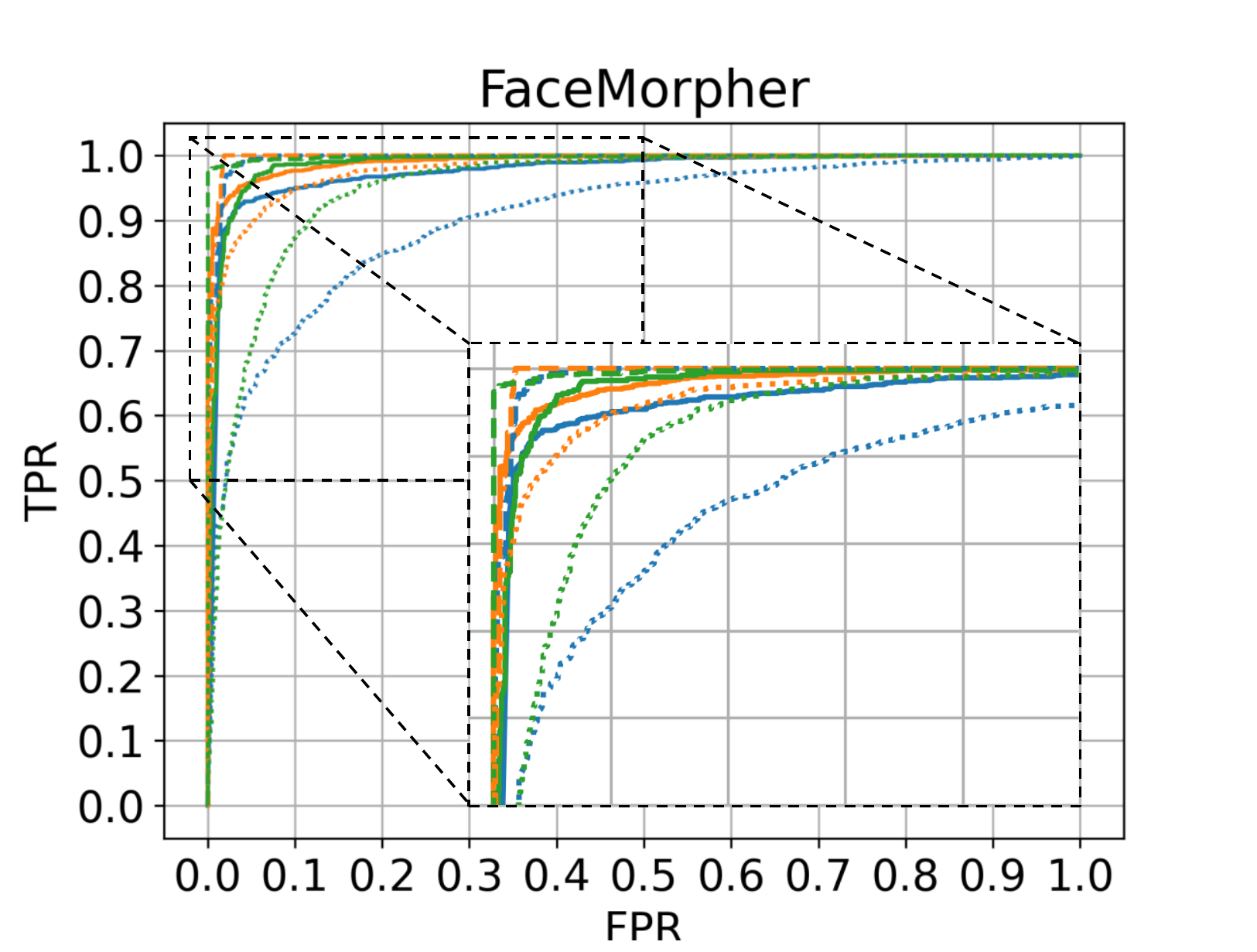}
  \hspace{-15 pt}
  \includegraphics[width=0.345\linewidth]{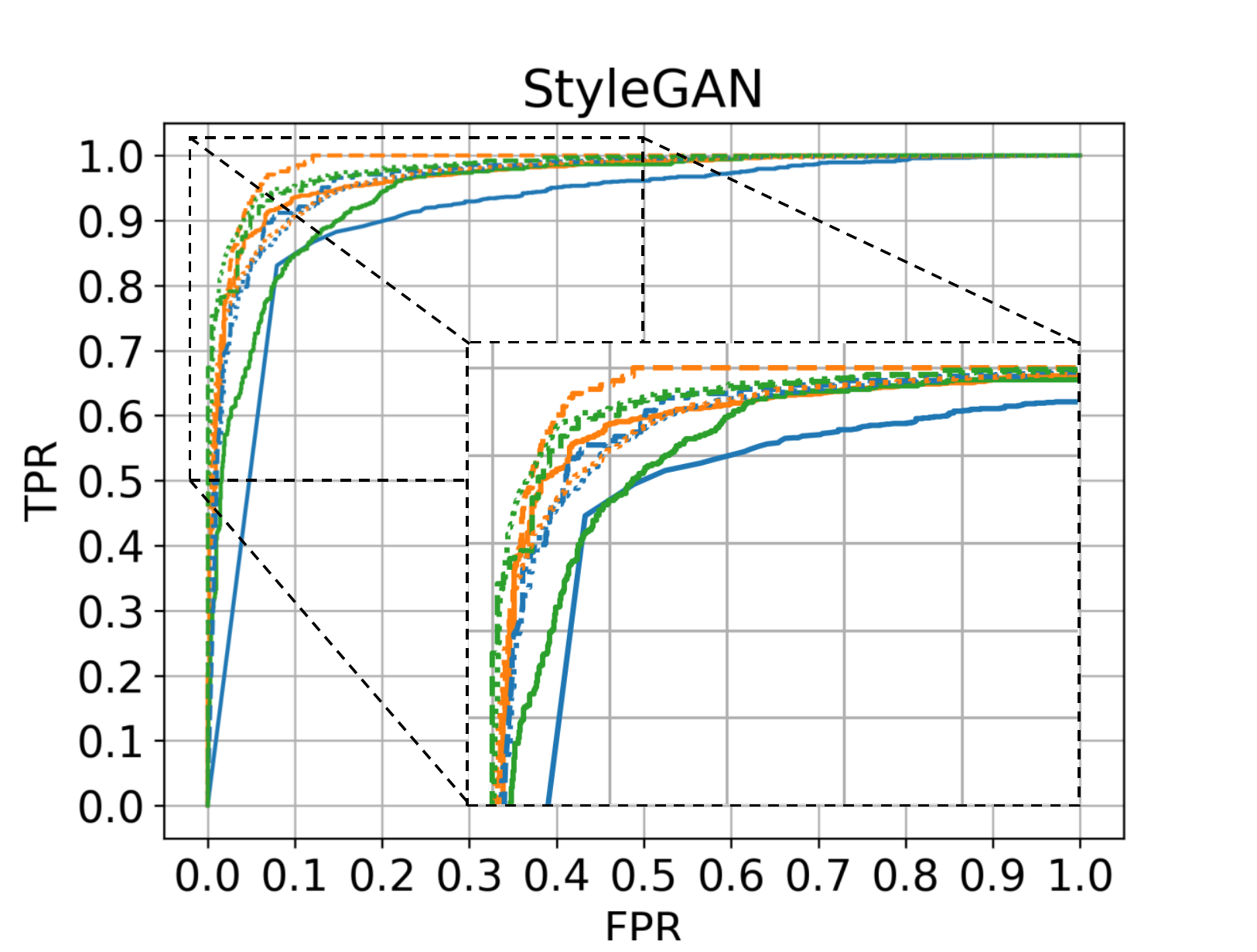}
  \includegraphics[width=0.345\linewidth]{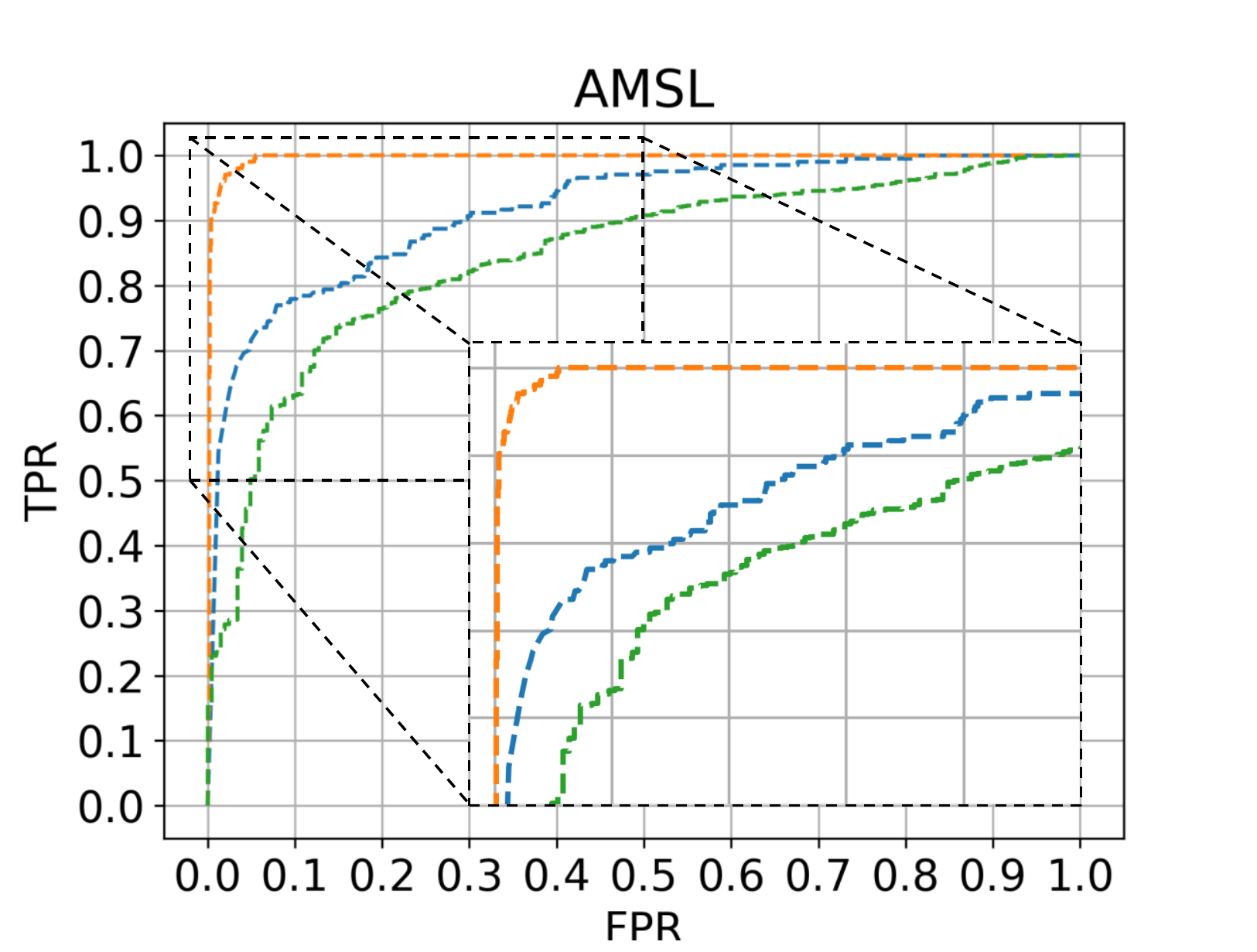}
  \hspace{-15 pt}
  \includegraphics[width=0.345\linewidth]{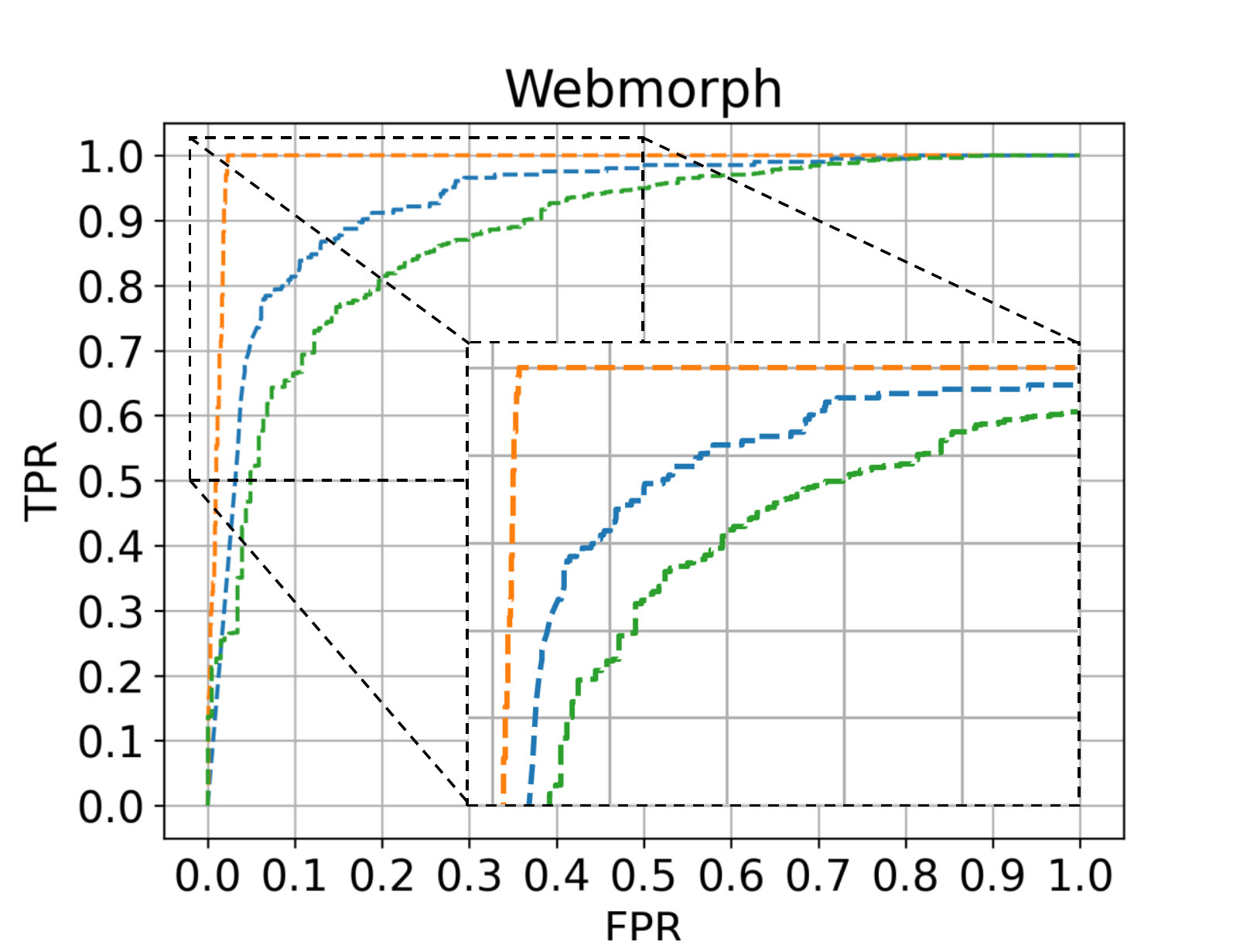}\hspace{-15 pt}
  \includegraphics[width=0.345\linewidth]{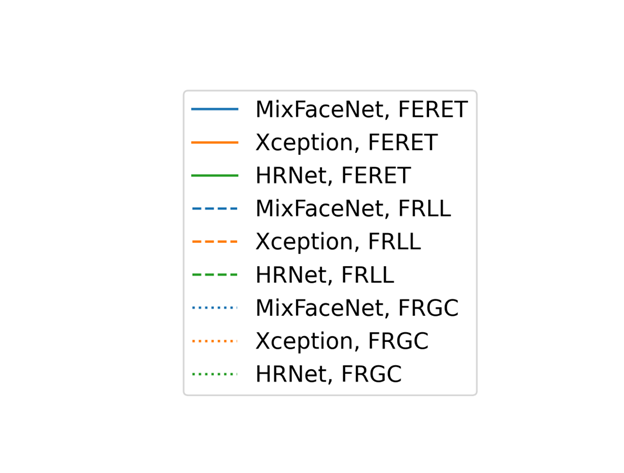}
  
\end{center}
\vspace{-20 pt}
\caption{ROC curves generated for the tested MAD models for different types of face morphs. Note that the performance of the detectors differs quite considerably depending on the morphing procedure used.\label{fig:ROC_by_morph_types}}
\end{figure*}

To better understand the differences between the evaluated models, we additionally assess their performance on only one face morphing technique at a time. As can be seen from Figure~\ref{fig:ROC_by_morph_types}, Xception shows the greatest generalization capabilities, when it comes to detection of different face morphing methods. HRNet provides very competitive results on face morphs generated by OpenCV, FaceMorpher and StyleGAN. Nevertheless, AMSL and Webmorph attacks are too challenging for this model. We hypothesize, that this might be due to the structure of the training data. Synthetic face morphs from SMDD are generated using only one morphing method, i.e. OpenCV. With such training data, models are in general at higher risk of overfitting to one specific type of morph. Since HRNet has far more trainable parameters than Xception and MixFaceNet, it is also more prone to overfitting, when trained on smaller datasets like SMDD.
\vspace{-3 pt}
\section{Conclusion}
\vspace{-3 pt}
In this paper, we tackle the privacy issues associated with the datasets used for the development of face morphing detection algorithms. To address related privacy concerns, we explore the idea of using a training database with faces of non--existing people, generated by StyleGAN. Using this data, we train three different MAD models and evaluate their performance on three commonly used real--world datasets. Our experiments show that in general, MAD models %based on out--of--the--box binary classifiers 
can be successfully trained on synthetic data and generalize well to real--world scenarios. %However, models somethimes fail to detect certain types of face morphing techniques, which were not utilized for the generation of the training data. Additional experiments are therefore needed to further analyze the correlation between the structure of the synthetic training dataset and the capacity of the learned detectors to learn useful features in the considered training settings. 

\footnotesize %
%\small
%\printbibliography
\bibliographystyle{ieee}
\bibliography{bibliography}

\begin{thebibliography}{10}\itemsep=-1pt

\bibitem{Boutros_MixFaceNet_2021}
F.~Boutros, N.~Damer, M.~Fang, F.~Kirchbuchner, and A.~Kuijper.
\newblock Mixfacenets: Extremely efficient face recognition networks.
\newblock In {\em IEEE IJCB}, pages 1--8, 2021.

\bibitem{Xception}
F.~Chollet.
\newblock {Xception: Deep Learning with Depthwise Separable Convolutions}.
\newblock In {\em IEEE CVPR}, pages 1800--1807, 2017.

\bibitem{Damer_SMDD_2022}
N.~Damer, C.~A.~F. L\'opez, M.~Fang, N.~Spiller, M.~V. Pham, and F.~Boutros.
\newblock {Privacy-Friendly Synthetic Data for the Development of Face Morphing
  Attack Detectors}.
\newblock {\em IEEE CVPRW}, pages 1606--1617, 2022.

\bibitem{Naser_PW_MAD_2021}
N.~Damer, N.~Spiller, M.~Fang, F.~Boutros, F.~Kirchbuchner, and A.~Kuijper.
\newblock {PW-MAD: Pixel-Wise Supervision for Generalized Face Morphing Attack
  Detection}.
\newblock In {\em Advances in Visual Computing}, pages 291--304. Springer
  International Publishing, 2021.

\bibitem{debruine_jones_2017}
L.~DeBruine and B.~Jones.
\newblock {Face Research Lab London Set}, 2017.

\bibitem{RetinaFace}
J.~Deng, J.~Guo, E.~Ververas, I.~Kotsia, and S.~Zafeiriou.
\newblock {RetinaFace: Single-Shot Multi-Level Face Localisation in the Wild}.
\newblock In {\em IEEE CVPR}, pages 5202--5211, 2020.

\bibitem{grm2018strengths}
K.~Grm, V.~{\v{S}}truc, A.~Artiges, M.~Caron, and H.~K. Ekenel.
\newblock {Strengths and weaknesses of deep learning models for face
  recognition against image degradations}.
\newblock {\em IET Biometrics}, 7(1):81--89, 2018.

\bibitem{Huber_SYN-MAD_2022}
M.~Huber, F.~Boutros, A.~Thi~Luu, K.~Raja, R.~Ramachandra, N.~Damer, P.~C.
  Neto, T.~Goncalves, A.~F. Sequeira, J.~S. Cardoso, T.~Joao, M.~Lourenc,
  S.~Serra, E.~Cermeno, M.~Ivanovska, B.~Batagelj, A.~Kronovsek, P.~Peer, and
  V.~Struc.
\newblock {SYN-MAD 2022: Competition on Face Morphing Attack Detection Based on
  Privacy-aware Synthetic Training Data}.
\newblock In {\em IEEE IJCB}, 2022.

\bibitem{Karras_stylegan2_2020}
T.~Karras, M.~Aittala, J.~Hellsten, S.~Laine, J.~Lehtinen, and T.~Aila.
\newblock {Training Generative Adversarial Networks with Limited Data}.
\newblock In {\em NIPS}, 2020.

\bibitem{karras2020analyzing}
T.~Karras, S.~Laine, M.~Aittala, J.~Hellsten, J.~Lehtinen, and T.~Aila.
\newblock {Analyzing and Improving the Image Quality of StyleGAN}.
\newblock In {\em IEEE/CVF CVPR}, pages 8110--8119, 2020.

\bibitem{Makrushin_SURF_2019}
A.~Makrushin, C.~Kraetzer, J.~Dittmann, C.~Seibold, A.~Hilsmann, and P.~Eisert.
\newblock {Dempster-Shafer Theory for Fusing Face Morphing Detectors}.
\newblock In {\em EUSIPCO}, pages 1--5, 2019.

\bibitem{Neubert_FRLL}
T.~Neubert, A.~Makrushin, M.~Hildebrandt, C.~Kraetzer, and J.~Dittmann.
\newblock {Extended StirTrace benchmarking of biometric and forensic qualities
  of morphed face images}.
\newblock {\em IET Biometrics}, 7(4):325--332, 2018.

\bibitem{Ojala_1996_LBP}
T.~Ojala, M.~Pietikäinen, and D.~Harwood.
\newblock A comparative study of texture measures with classification based on
  featured distributions.
\newblock {\em Pattern Recognition}, 29(1):51--59, 1996.

\bibitem{Ojansivu_LPQ_2008}
V.~Ojansivu and J.~Heikkil{\"a}.
\newblock {Blur Insensitive Texture Classification Using Local Phase
  Quantization}.
\newblock In A.~Elmoataz, O.~Lezoray, F.~Nouboud, and D.~Mammass, editors, {\em
  Image and Signal Processing}, pages 236--243. Springer Berlin Heidelberg,
  2008.

\bibitem{Phillips_FRGC}
P.~Phillips, P.~Flynn, T.~Scruggs, K.~Bowyer, J.~Chang, K.~Hoffman, J.~Marques,
  J.~Min, and W.~Worek.
\newblock Overview of the face recognition grand challenge.
\newblock In {\em IEEE CVPR}, volume~1, pages 947--954 vol. 1, 2005.

\bibitem{PHILLIPS_FERET}
P.~J. Phillips, H.~Wechsler, J.~Huang, and P.~J. Rauss.
\newblock {The FERET database and evaluation procedure for face-recognition
  algorithms}.
\newblock {\em Image and Vision Computing}, 16(5):295--306, 1998.

\bibitem{Raghavendra_transfer_2017}
R.~Raghavendra, K.~B. Raja, S.~Venkatesh, and C.~Busch.
\newblock {Transferable Deep-CNN Features for Detecting Digital and
  Print-Scanned Morphed Face Images}.
\newblock In {\em IEEE CVPRW}, pages 1822--1830, 2017.

\bibitem{Ramachandra_Inception_2020}
R.~Ramachandra, S.~Venkatesh, K.~Raja, and C.~Busch.
\newblock {Detecting Face Morphing Attacks with Collaborative Representation of
  Steerable Features}.
\newblock In {\em CVIP}, pages 255--265, 2020.

\bibitem{Sarkar2020_morphed_data}
E.~Sarkar, P.~Korshunov, L.~Colbois, and S.~Marcel.
\newblock {Vulnerability Analysis of Face Morphing Attacks from Landmarks and
  Generative Adversarial Networks}.
\newblock 2020.

\bibitem{Scherhag_PRNU_2019}
U.~Scherhag, L.~Debiasi, C.~Rathgeb, C.~Busch, and A.~Uhl.
\newblock {Detection of Face Morphing Attacks Based on PRNU Analysis}.
\newblock {\em IEEE TBBIS}, 1(4):302--317, 2019.

\bibitem{Seibold_reflection_2018}
C.~Seibold, A.~Hilsmann, and P.~Eisert.
\newblock {Reflection Analysis for Face Morphing Attack Detection}.
\newblock In {\em EUSIPCO}, pages 1022--1026, 2018.

\bibitem{szegedy_rethinking_2015}
C.~Szegedy, V.~Vanhoucke, S.~Ioffe, J.~Shlens, and Z.~Wojna.
\newblock Rethinking the {Inception} {Architecture} for {Computer} {Vision}.
\newblock {\em CoRR}, 2015.

\bibitem{Wandzik_FRS_2018}
L.~Wandzik, G.~Kaeding, and R.~V. Garcia.
\newblock {Morphing Detection Using a General-Purpose Face Recognition System}.
\newblock In {\em EUSIPCO}, pages 1012--1016, 2018.

\bibitem{Wang_HRNet_2019}
J.~Wang, K.~Sun, T.~Cheng, B.~Jiang, C.~Deng, Y.~Zhao, D.~Liu, Y.~Mu, M.~Tan,
  X.~Wang, W.~Liu, and B.~Xiao.
\newblock {Deep High-Resolution Representation Learning for Visual
  Recognition}.
\newblock {\em TPAMI}, 2019.

\end{thebibliography}

\end{document}